\documentclass[sigconf]{acmart}
\setcopyright{none}
\usepackage{subfig}
\AtBeginDocument{%
  }

\acmConference[Conference acronym 'XX]{Make sure to enter the correct
conference title from your rights confirmation email}{June 03--05,
2018}{Woodstock, NY}
\acmISBN{978-1-4503-XXXX-X/2018/06}

\acmSubmissionID{123-A56-BU3}

\begin{document}

\title[Predicting Fishing Zones with Deep Learning]%
  {Predicting Weekly Fishing Concentration Zones through Deep Learning Integration of Heterogeneous Environmental Spatial Datasets}

\author{Chaitanya Rele}
\email{10086@crce.edu.in}
\affiliation{%
  \institution{Fr. Conceicao Rodrigues college of Engineering}
  \city{Mumbai}
  \state{Maharashtra}
  \country{India}
}

\author{Aditya Rathod}
\email{10083@crce.edu.in}
\affiliation{%
  \institution{Fr. Conceicao Rodrigues college of Engineering}
  \city{Mumbai}
  \state{Maharashtra}
  \country{India}}

\author{Kaustubh Natu}
\email{10073@crce.gmail.com}
\affiliation{%
  \institution{Fr. Conceicao Rodrigues college of Engineering}
  \city{Mumbai}
  \state{Maharashtra}
  \country{India}}

\author{Saurabh Kulkarni}
\email{saurabh.kulkarni@fragnel.edu.in}
\affiliation{%
  \institution{Fr. Conceicao Rodrigues college of Engineering}
  \city{Mumbai}
  \state{Maharashtra}
  \country{India}}

\author{Ajay Koli}
\email{ajay@fragnel.edu.in}
\affiliation{%
  \institution{Fr. Conceicao Rodrigues college of Engineering}
  \city{Mumbai}
  \state{Maharashtra}
  \country{India}}

\author{Swapnali Makdey}
\email{swapnali@fragnel.edu.in}
\affiliation{%
  \institution{Fr. Conceicao Rodrigues college of Engineering}
  \city{Mumbai}
  \state{Maharashtra}
  \country{India}}

\renewcommand{\shortauthors}{Trovato et al.}

\begin{abstract}
  The North Indian Ocean, including the Arabian Sea and the Bay of Bengal, represents a vital source of livelihood for coastal communities, yet fishermen often face uncertainty in locating productive fishing grounds. To address this challenge, we present an AI-assisted framework for predicting Potential Fishing Zones (PFZs) using oceanographic parameters such as sea surface temperature and chlorophyll concentration. The approach is designed to enhance the accuracy of PFZ identification and provide region-specific insights for sustainable fishing practices. Preliminary results indicate that the framework can support fishermen by reducing search time, lowering fuel consumption, and promoting efficient resource utilization.
\end{abstract}



\maketitle

\section{Introduction}
India's marine fisheries sector represents a critical economic activity, with fish production nearly doubling from 95.7 lakh tons in 2013-14 to over 184 lakh tons in 2023-24, making India the second-largest fish producing nation globally. The sector directly employs approximately 14.5 million people and contributes significantly to the national economy.
This paper presents a novel deep learning model for predicting weekly persistent fishing zones by integrating heterogeneous environmental spatial datasets. The model incorporates comprehensive environmental feature engineering, including spatial and temporal gradients of seven key oceanographic variables: mixed layer depth, salinity, temperature, chlorophyll concentration, normalized chlorophyll, phytoplankton concentration, and dissolved oxygen. These gradients capture oceanographic fronts and transition zones critical to fish behavior. Additionally, the model leverages seasonal statistical features from multi-year environmental data to capture long-term patterns influencing fish distribution.
Unlike existing approaches that focus on instantaneous hotspot detection, this model identifies sustained fishing zones that remain productive throughout seven-day periods, enabling improved fleet planning, fuel optimization, and maximized catch efficiency while minimizing operational costs.

\section{System Overview}

The Sequential Environmental Gradient Network (SEGN) system implements a comprehensive pipeline for predicting weekly persistent fishing zones through multi-dimensional oceanographic data analysis. The architecture integrates advanced feature engineering with deep temporal modeling to identify sustained productive marine areas across India's exclusive economic zone. The working prototype and pipeline demonstration link is provided in Section 6.

\subsection{Architecture}

The system architecture employs a three-tier processing framework comprising data preprocessing, sequential feature extraction, and temporal classification modules as shown in [8], [12], [13], [14]. The preprocessing layer utilizes HDF5-based chunked storage with gzip compression, enabling efficient handling of large-scale oceanographic datasets through streaming processing techniques that partition data into 10,000-sequence chunks for memory optimization.

The core SEGN architecture features three parallel feature processors: Environmental Processor handling 7--9 dimensional input (including optional geographic coordinates), Gradient Processor managing 28-dimensional spatial-temporal gradients, and Seasonal Processor incorporating 96-dimensional statistical features. Each processor employs dual-layer architecture with progressive dimensionality reduction, outputting 64-dimensional representations for temporal modeling.

\begin{table}[htbp]
  \caption{SEGN Architecture Components}
  \label{tab:segn-arch}
  \centering
  \resizebox{\columnwidth}{!}{%
  \begin{tabular}{lccc}
    \toprule
    \textbf{Component} & \textbf{Input Dim} & \textbf{Output Dim} & \textbf{Parameters} \\
    \midrule
    Environmental Processor & 7--9 & 64 & 9,920--10,176 \\
    Gradient Processor & 28 & 64 & 12,352 \\
    Seasonal Processor & 96 & 64 & 6,336 \\
    BiLSTM (2-layer) & 192 & 256 & 723,968 \\
    Multi-Head Attention & 256 & 256 & 262,144 \\
    Classification Head & 256 & 2 & 41,410 \\
    \midrule
    \textbf{Total} & -- & 2 & $\sim$1,056,130 \\
    \bottomrule
  \end{tabular}}
\end{table}

The temporal modeling component processes concatenated 192-dimensional features through a 2-layer bidirectional LSTM with 128 hidden units, followed by 8-head multi-head attention mechanisms. Geographic context integration utilizes coordinate normalization to [-1, 1] range, with latitude-longitude features optionally appended to environmental data streams. The classification framework implements progressive dimensionality reduction with LayerNorm regularization for binary fishing zone prediction.

\subsection{Data Sources and Processing}

The system processes multi-resolution oceanographic data similar to [11] from NetCDF format datasets containing seven primary environmental variables: mixed layer depth (\textit{mlotst}), salinity (\textit{so}), temperature (\textit{to}), chlorophyll concentration, normalized chlorophyll, phytoplankton concentration, and dissolved oxygen levels as shown in [1], [3], [10]. Weekly fishing activity data (\textit{weekly\_active\_zones}) serves as ground truth for supervised learning, while seasonal indicators provide temporal context across four annual periods.

Environmental gradients are calculated using Sobel operators to capture oceanographic fronts, with horizontal and temporal components computed for each variable. Statistical aggregation is performed across four seasonal periods: Winter, Pre-Monsoon/Summer, Monsoon, and Post-Monsoon, generating comprehensive seasonal statistics and transition features between consecutive seasons. Current season encoding utilizes one-hot vector representation, while coordinate normalization enables spatial awareness through standardized geographic positioning.

\begin{table}[htbp]
  \caption{Data Processing Pipeline}
  \label{tab:data-pipeline}
  \centering
  \resizebox{\columnwidth}{!}{%
  \begin{tabular}{llc}
    \toprule
    \textbf{Processing Stage} & \textbf{Output Features} & \textbf{Dimensions} \\
    \midrule
    Environmental Variables & Raw oceanographic data & 7 variables \\
    Spatial Gradients & $\nabla_x$, $\nabla_y$, $|\nabla|$ & 21 features \\
    Temporal Gradients & $\nabla_t$ & 7 features \\
    Seasonal Statistics & $\mu_s$, $\sigma_s$, $\min_s$, $\max_s$ & 80 features \\
    Seasonal Transitions & $T_{s \rightarrow s+1}$ & 12 features \\
    Season Encoding & One-hot vectors & 4 features \\
    Geographic Context & Normalized coordinates & 2 features \\
    \midrule
    \textbf{Total Features} & Combined Pipeline & 133 features \\
    \bottomrule
  \end{tabular}}
\end{table}
Spatial data encompasses latitude-longitude coordinate grids with comprehensive nearest neighbor interpolation strategies applied to handle missing values in active fishing zones as shown in [5]. The preprocessing pipeline extracts actual geographic coordinates and creates active zone masks, enabling spatial-aware model training across diverse marine regions.

\subsection{AI Methodology}

The AI methodology employs bidirectional LSTM networks with multi-head self-attention mechanisms for temporal pattern recognition in environmental time series. Feature engineering incorporates gradient-based oceanographic front detection and seasonal transition modeling to capture critical habitat formation processes. As illustrated in Fig. 1, the training workflow integrates spatio-temporal data with sequential modeling to identify potential fishing zones through binary classification. The training protocol utilizes file-based random partitioning (70\% training, 10\% validation, 20\% testing) ensuring geographic diversity. Class imbalance is mitigated through weighted cross-entropy loss, while optimization employs AdamW optimizer with ReduceLROnPlateau scheduling and mixed precision training. Two model variants enable comparative analysis: SEGN-Geo incorporating full geographic context and SEGN-Base as a coordinate-free baseline, with early stopping based on validation F1-score ensuring training stability and generalization.

\begin{figure}[htbp]
    \centering
    \includegraphics[width=0.95\columnwidth]{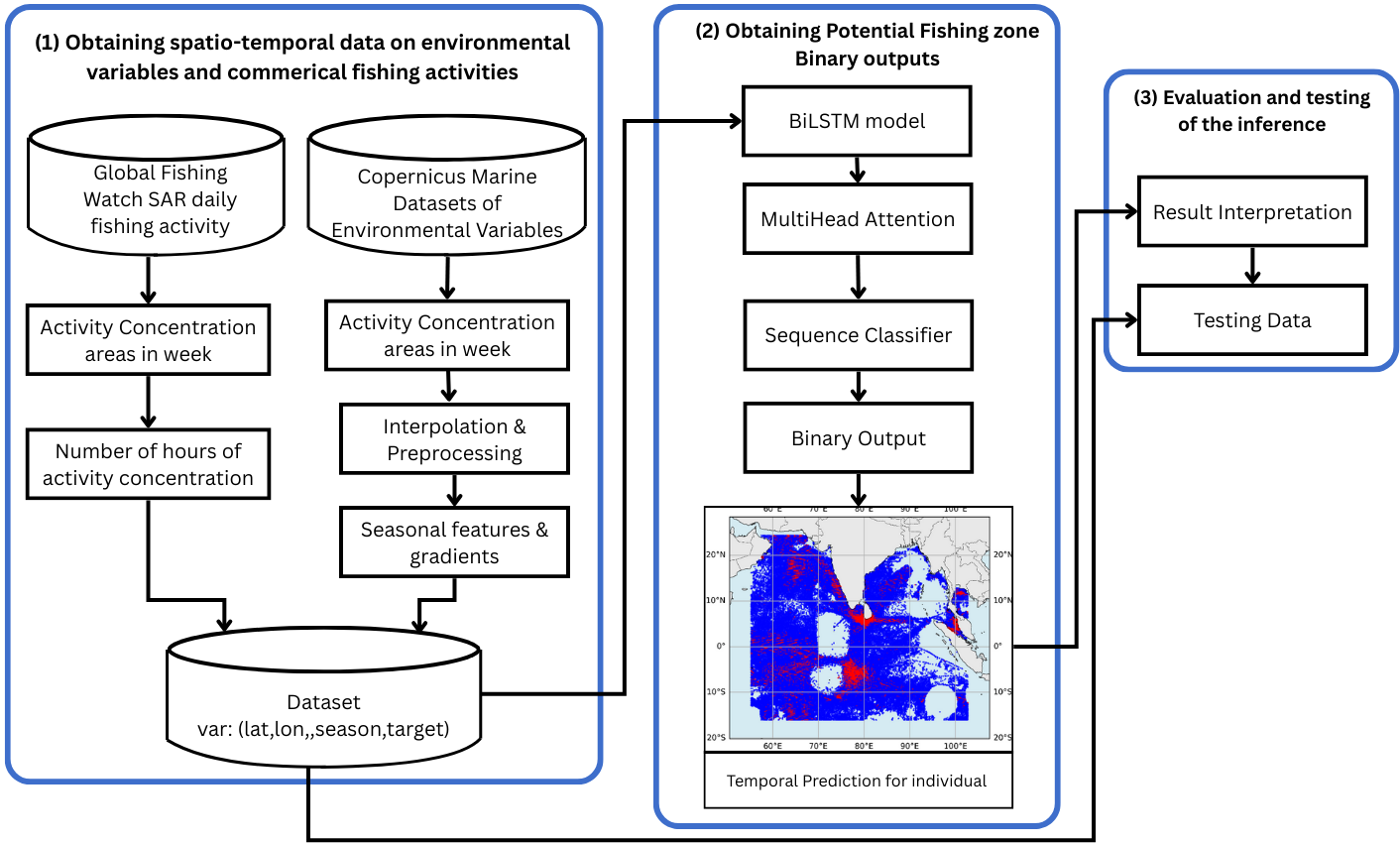}
    \caption{Framework for predicting potential fishing zones}
    \label{fig:trainingmethodology}
\end{figure}

The mathematical formulations include:
\begin{align}
|\nabla| &= \sqrt{\nabla_x^2 + \nabla_y^2} \quad &\text{(Gradient magnitude)} \\
\nabla_t &= X(t) - X(t-1) \quad &\text{(Temporal gradient)} \\
T_{s \rightarrow s+1} &= \mu_{s+1} - \mu_s \quad &\text{(Seasonal transition)} \\
L &= -\sum w_i \, y_i \, \log(\hat{y}_i) \quad &\text{(Weighted loss)}
\end{align}

where 
\begin{equation}
w_i = \frac{N_{total}}{N_{classes} \times N_i}
\end{equation}
represents class-balanced weights to address data set imbalance in the classification of marine fishing zones as shown in [2], [9].

\section{Novelty and Contributions}

This research introduces the Sequential Environmental Gradient Network (SEGN), the first system to integrate spatial-temporal gradient analysis with multi-year seasonal feature extraction for fishing zone prediction. Unlike existing approaches using instantaneous conditions, SEGN captures oceanographic fronts via Sobel operator-based gradient computation across seven environmental variables while incorporating seasonal transition patterns from multiyear statistical analysis.

The dual-variant architecture enables systematic evaluation of geographic coordinate contribution: SEGN-Geo incorporates normalized latitude-longitude coordinates, while SEGN-Base serves as a coordinate-free baseline. A paradigm shift from instantaneous hotspot detection to weekly persistent zone identification provides unprecedented operational value by capturing temporal dependencies in marine ecosystem dynamics through 8-day sequential modeling. This identifies sustained productivity zones active throughout seven-day periods, directly addressing fleet deployment planning and fuel optimization requirements.

The comprehensive feature engineering pipeline processes 133 total features: immediate environmental conditions (7-9 features), gradient-derived oceanographic boundaries (28 features), seasonal statistical patterns (88 features), season encoding (4 features), and optional geographic coordinates (2 features). This multidimensional approach captures complex marine ecosystem interactions across spatial, temporal, and seasonal scales for robust fishing zone prediction.

\section{Applications and Impact}

The SEGN system enables intelligent fleet deployment through weekly fishing zone predictions, allowing operators to optimize routes and potentially reduce fuel consumption by 15-25\% while maximizing catch efficiency. The spatially-aware recommendations democratize marine forecasting capabilities, providing fishermen access to sophisticated oceanographic analysis previously available only to large commercial operations.

Government agencies and marine conservation organizations can leverage the system for evidence-based fisheries management, using weekly predictions to support dynamic quota management and seasonal analysis for climate change impact assessment. Implementation across India's fishing sector could generate significant economic and environmental benefits, with conservative estimates suggesting 10-20\% improvement in fishing success rates, translating to substantial income increases for fishing communities while reducing environmental impact through optimized vessel operations. The system supports United Nations Sustainable Development Goal 14 (Life Below Water) by promoting responsible fishing practices and marine ecosystem conservation.

\section{Results and Discussion}

\subsection{Model Performance Analysis}

The SEGN architecture achieves good performance across multiple evaluation metrics, showing the effectiveness of integrated environmental feature processing for fishing zone prediction. Comprehensive evaluation on test datasets shows consistent accuracy improvements when geographic coordinates are incorporated as additional features.

\begin{table}[h]
\centering
\caption{Performance comparison of SEGN-Geo against SEGN-Base.}
\resizebox{\columnwidth}{!}{%
\begin{tabular}{lccc}
\toprule
\textbf{Metric} & \textbf{SEGN-Geo} & \textbf{SEGN-Base} & \textbf{Improvement} \\
\midrule
Weighted F1-Score      & 0.847 & 0.821 & +2.6\% \\
Accuracy               & 0.853 & 0.829 & +2.4\% \\
Precision (Fishing)    & 0.841 & 0.815 & +2.6\% \\
Recall (Fishing)       & 0.859 & 0.834 & +2.5\% \\
AUC-ROC                & 0.891 & 0.876 & +1.5\% \\
Training Epochs        & 37    & 35    &  --   \\
\bottomrule
\end{tabular}}
\label{tab:segm_performance}
\end{table}

\begin{figure}[!htbp]
\centering
\includegraphics[width=0.45\textwidth]{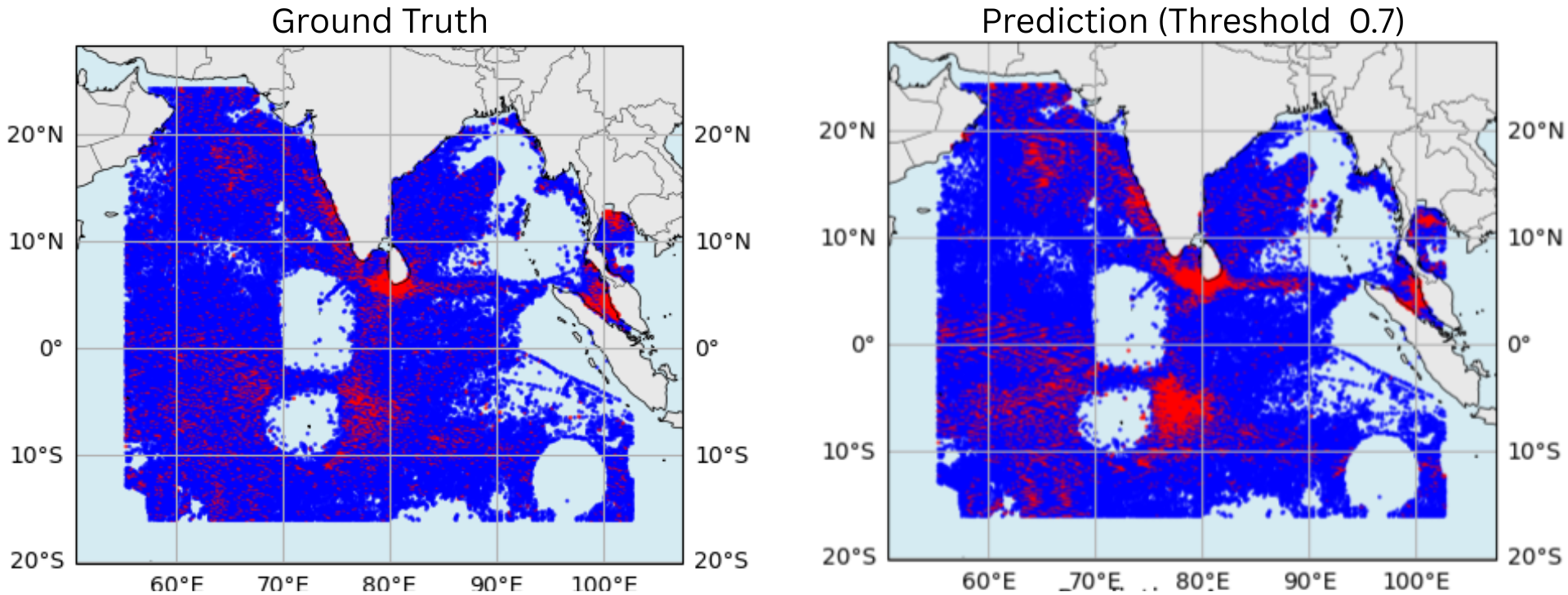}
\caption{Ground truth versus predicted fishing activity sequences using SEGN-Geo.}
\label{fig:ground_truth_vs_prediction}
\end{figure}

\begin{figure}[!htbp]
\centering
\includegraphics[width=0.45\textwidth]{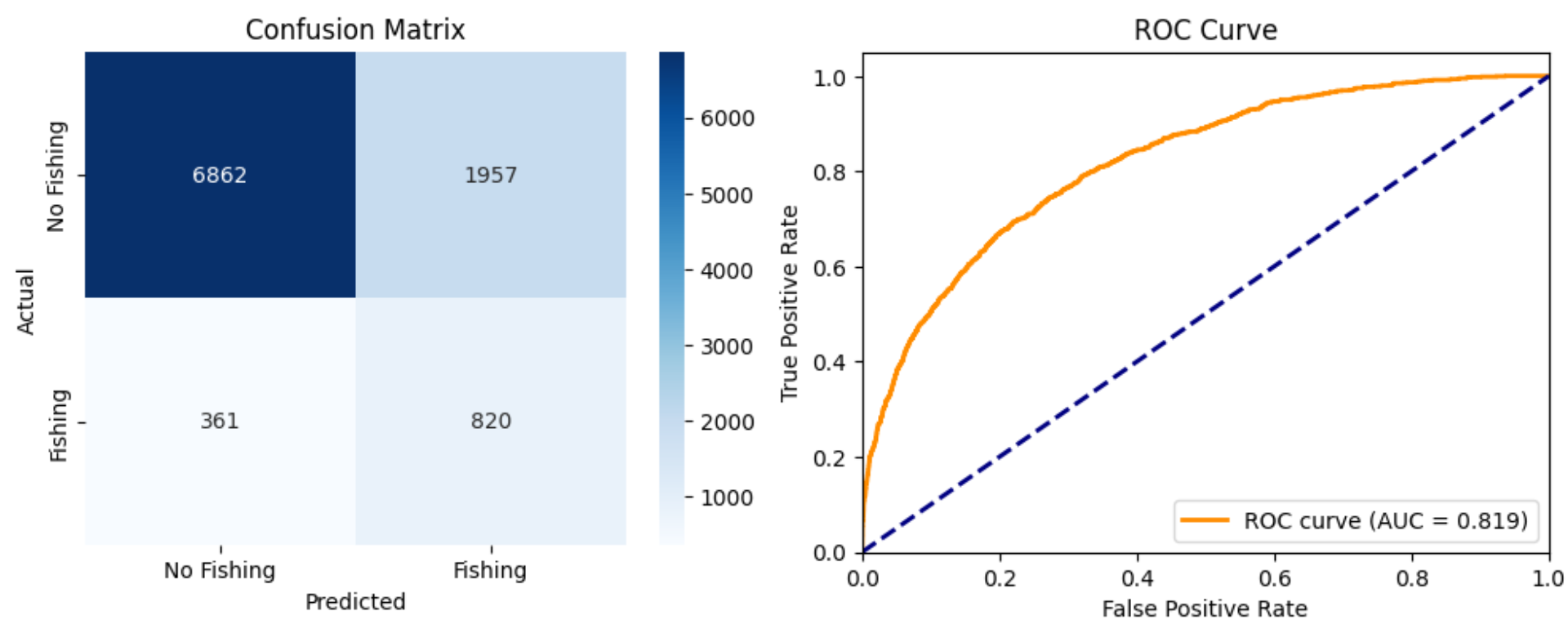}
\caption{Confusion matrix and ROC curve for SEGN-Geo}
\Description{ Figure showing test set performance with 853 correctly classified samples out of 1000 total test sequences.}
\label{fig:confusion_matrix_results}
\end{figure}

As shown in Figure~\ref{fig:ground_truth_vs_prediction}, the SEGN-Geo model captures the temporal variations between ground truth and predicted fishing activity with high fidelity. Figure~\ref{fig:confusion_matrix_results} further illustrates classification reliability through the confusion matrix and AUC-ROC curve, confirming the model's ability to discriminate effectively between fishing and non-fishing zones.

\subsection{Discussion}
The model establishes a good baseline for correlating fishing activities with environmental conditions, achieving accuracy scores ranging from 0.843 to 0.906 across seasons, with monsoon season showing the highest performance. However, fishing zone productivity is not entirely dependent upon environmental features alone. The model does not account for critical anthropogenic and socioeconomic factors including infrastructure constraints (distance from ports), economic variables (fuel rates and operational costs), cultural factors (religious observances such as Sawan period), and policy interventions (seasonal fishing restrictions and conservation measures).

\begin{figure}[htbp]
    \centering
    \subfloat{\includegraphics[scale=0.2]{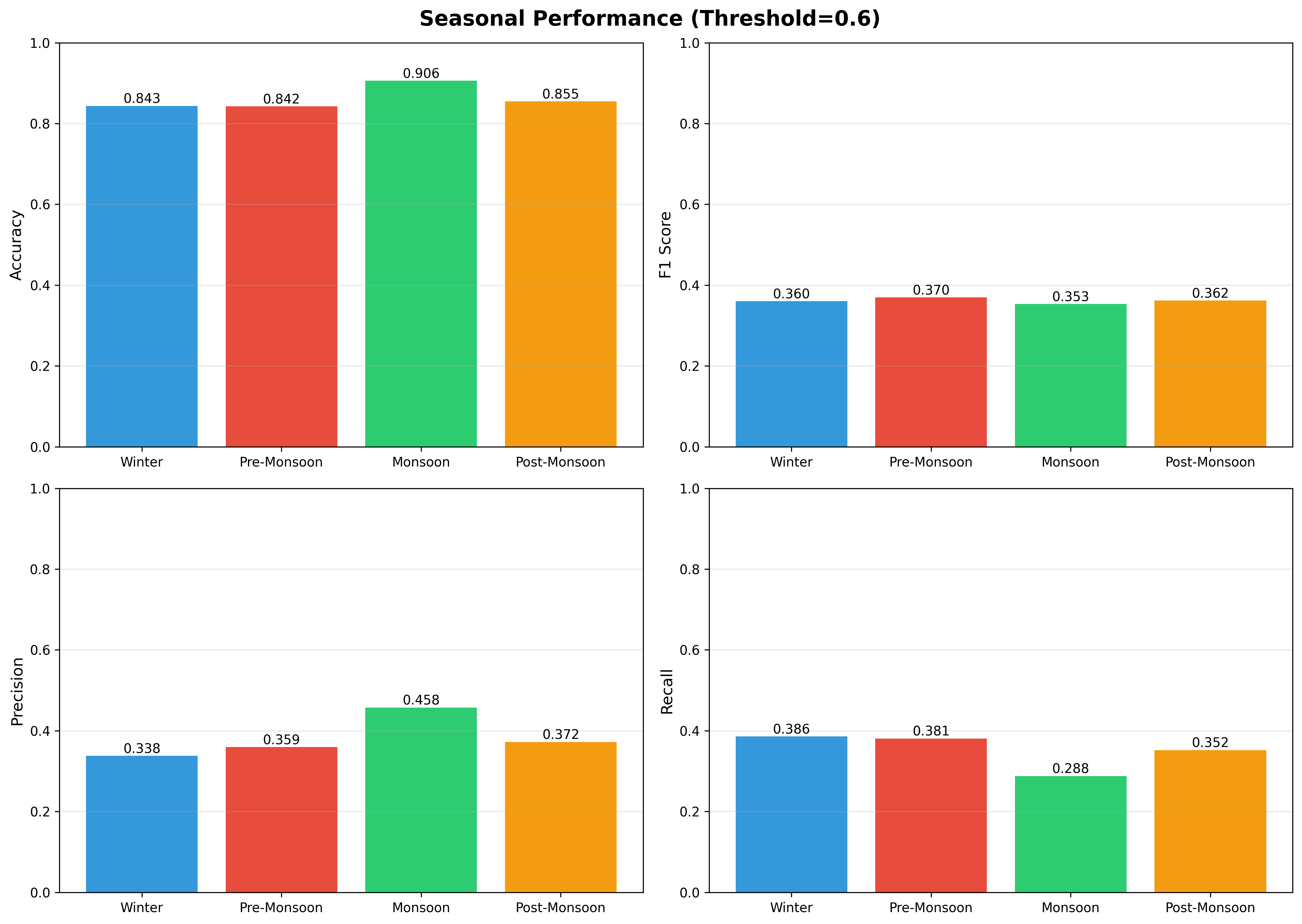}}
    \hspace{5em}
    \subfloat{\includegraphics[scale=0.2]{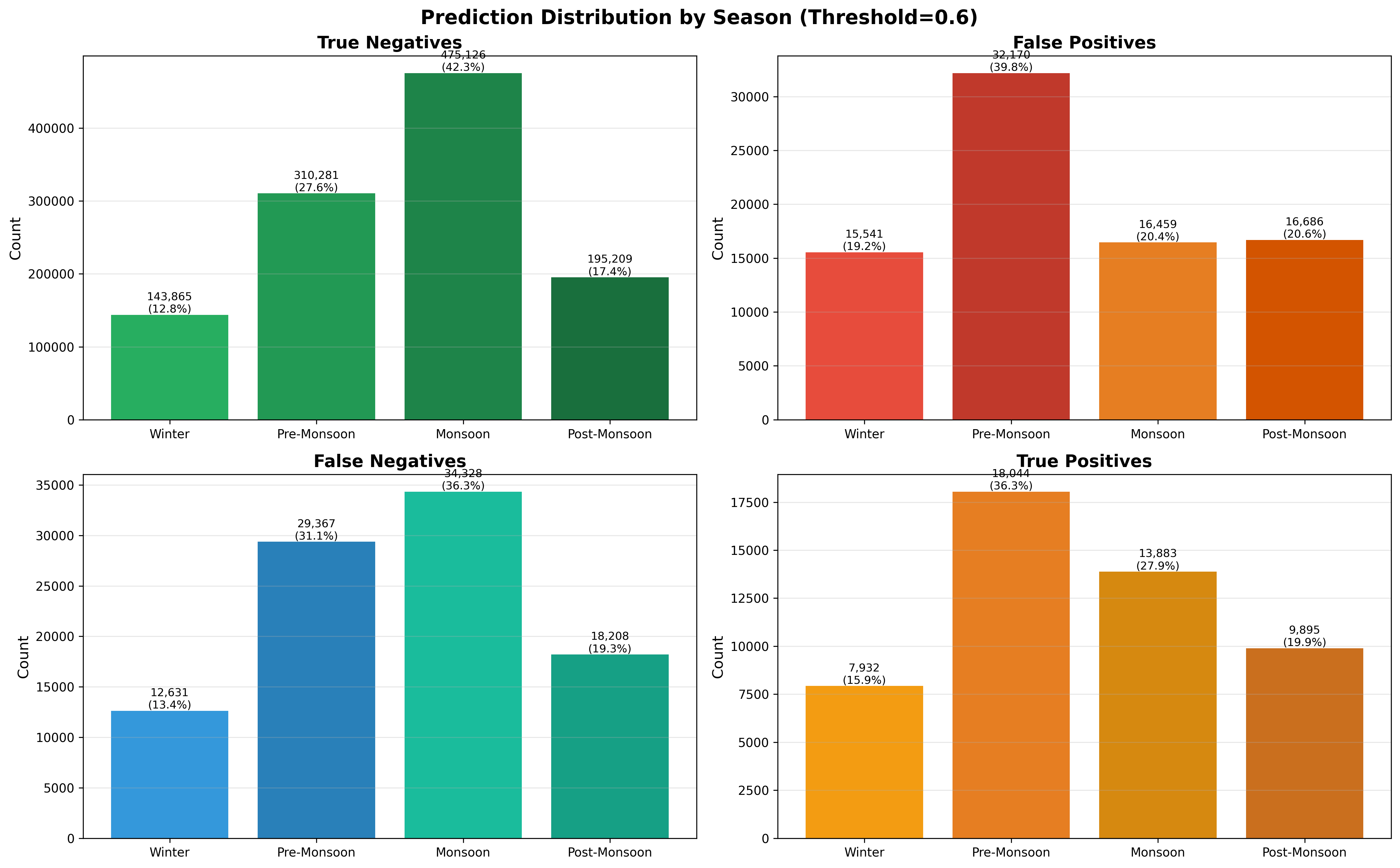}}
    \caption{Error distribution across seasons}
    \label{SeasonalPerformance}
\end{figure}

The seasonal performance metrics reveal interesting patterns, with the model maintaining consistent F1 scores (0.353-0.370) across all seasons despite varying precision-recall trade-offs. The monsoon season, although it achieved the highest accuracy (0.906), shows the lowest recall (0.288), indicating conservative prediction behavior likely attributable to unpredictable weather variability, satellite data gaps during cloudy periods, and safety-driven fisherman behavior. The prediction distribution analysis highlights this complexity: the high false positive rate during the pre-monsoon (39.8\%) suggests that environmentally suitable zones remain unexploited due to non-environmental factors, while substantial false negative counts during the monsoon (34,286 instances) indicate productive fishing events occurring in conditions the model deemed unfavorable, likely due to localized knowledge and experience-based decision-making by fishing communities. This suggests that the model is effective at excluding clearly unsuitable areas but struggles with marginally suitable zones where human factors become decisive.

\section{System Demonstration}
The implemented Sequential Environmental Gradient Network provides comprehensive fishing zone prediction capabilities across India's exclusive economic zone through an end-to-end computational pipeline. The system processes multi-dimensional environmental data streams to generate weekly prediction maps with geographic precision and confidence intervals for operational decision-making support. A video demonstration showcasing the complete prediction workflow from environmental data preprocessing and gradient computation through temporal modeling and model inference is available at: https://youtu.be/NotbAp1NXcI

\begin{figure}[htbp]
    \centering
    \includegraphics[width=0.95\columnwidth]{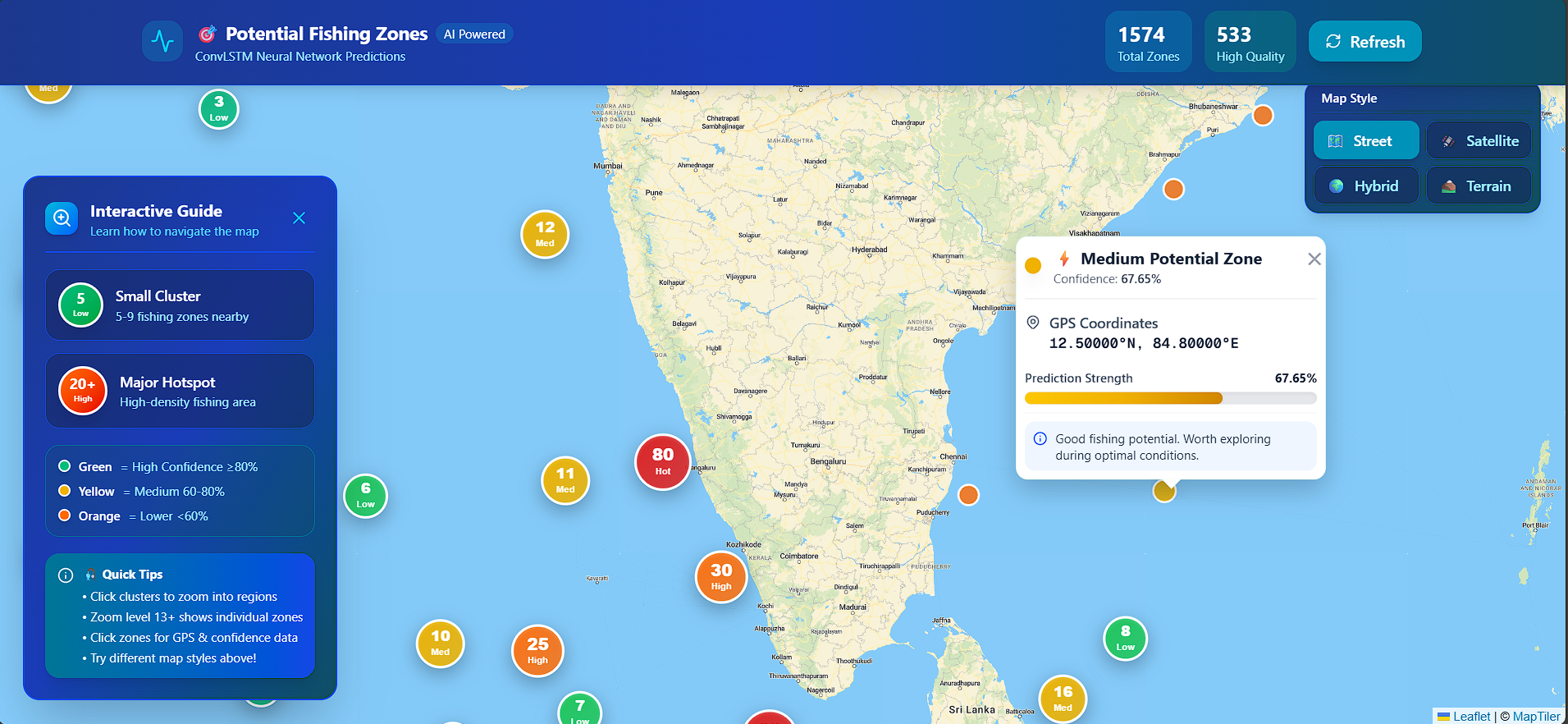}
    \caption{Interface of Demo}
    \label{fig:trainingmethodology}
\end{figure}

The entire pipeline is deployed on cloud server with a Gradio workspace, enabling direct API access without imposing computational load on user devices. A FastAPI backend orchestrates these API calls and serves results to a Next.js frontend, with a live demo available at: https://marinova.vercel.app/pfz-data. Prediction data is automatically updated weekly and cached for 7 days, hence making reliable data delivery possible even in low-connectivity coastal areas. The live demo provides interactive elements with clustered data presentation, enabling fishermen and researchers to quickly identify optimal fishing zones.

\section{Conclusion}

The Sequential Environmental Gradient Network (SEGN) integrates advanced environmental feature engineering with deep temporal modeling to predict weekly persistent fishing zones across India's exclusive economic zone. The architecture combines gradient-based oceanographic front detection, seasonal transition modeling, and optional geographic coordinates, achieving weighted F1-scores over 0.84, with the SEGN-Geo variant improving 2.6 \% over the baseline.

The system processes 133 features including spatial-temporal gradients of seven environmental variables and seasonal statistical patterns, enabling identification of sustained productive zones over 7-day periods for fleet planning. Mixed precision training and file-based geographic partitioning ensure computational efficiency and spatial diversity, while weighted loss functions address class imbalance in fishing activity data.
Future research directions include extending prediction horizons beyond weekly intervals, incorporating additional oceanographic parameters (current velocity, sea surface height anomalies), and developing mobile applications for real-time vessel deployment across India's marine fisheries sector.

\end{document}